%% file: main.tex
\documentclass[10pt,twocolumn,letterpaper]{article}

\usepackage[applications]{wacv}      % To produce the REVIEW version for the algorithms track

\definecolor{wacvblue}{rgb}{0.21,0.49,0.74}
\usepackage[pagebackref,breaklinks,colorlinks,allcolors=wacvblue]{hyperref}
\usepackage[table]{xcolor}
\usepackage{enumitem}
\usepackage{amsthm}
\newtheorem{definition}{Definition}
\usepackage[dvipsnames]{xcolor} % optional: loads many named colors
\usepackage{pifont}
\usepackage[utf8]{inputenc}

\usepackage[most]{tcolorbox}
\usepackage{fvextra} % for line breaking in code blocks
\tcbset{
  promptbox/.style={
    colback=gray!10,
    colframe=gray!50,
    boxrule=0.5pt,
    arc=3pt,
    left=4pt,
    right=4pt,
    top=4pt,
    bottom=4pt,
    fonttitle=\bfseries,
    enhanced,
    breakable,
  }
}

\def\wacvPaperID{767} % *** Enter the WACV Paper ID here
\def\confName{WACV}
\def\confYear{2026}

\title{Generalizing Sports Feedback Generation by Watching Competitions and Reading Books: A Rock Climbing Case Study}

\author{Arushi Rai\\
University of Pittsburgh\\
{\tt\small arr159@pitt.edu}
\and
Adriana Kovashka\\
University of Pittsburgh\\
{\tt\small kovashka@cs.pitt.edu}
}

\begin{document}
\maketitle
\begin{abstract}
%Athletes see incredible performance gains when a coach tells them relevant information about their performance and how to improve. Since it is not practical to always have a coach while practicing, and most people cannot afford a coach, there is a need for automatic sports feedback generation from videos.  
While there is rapid progress in video-LLMs with advanced reasoning capabilities, prior work shows that these models struggle on the challenging task of sports feedback generation and require expensive and difficult-to-collect finetuning feedback data for each sport. This limitation is evident from the poor generalization to sports unseen during finetuning. Furthermore, traditional text generation evaluation metrics (e.g., BLEU-4, METEOR, ROUGE-L, BERTScore), originally developed for machine translation and summarization, fail to capture the unique aspects of sports feedback quality. 
To address the first problem, using rock climbing as our case study, we propose using auxiliary freely-available web data from the target domain, such as competition videos and coaching manuals, in addition to existing sports feedback from a disjoint, source domain to improve sports feedback generation performance on the target domain. 
To improve evaluation, we propose two evaluation metrics: (1) specificity and (2) actionability. 
Together, our approach enables more meaningful and practical generation of sports feedback under limited annotations.
\end{abstract}
\input{sections/introduction}
\input{sections/related_works}
\input{sections/dataset}
\input{sections/method_rewrite}
\input{sections/experiments}
\input{sections/conclusion}
{
    \small
    \bibliographystyle{ieeenat_fullname}
    \bibliography{main}
}

\input{sections/supplemental}

\end{document}

%% file: sections/introduction.tex
\section{Introduction}

Athletes improve their performance through personalized coaching sessions where coaches provide specific insights on performance. However, a coach is not always available during self-directed practice sessions, and could be inaccessible financially or physically. This can limit performance improvement and satisfaction \cite{Losch2016}, both of which impact long-term motivation. Using AI to automatically generate feedback for athletic performances can make performance improvement accessible to all and at any time \cite{Seino2025ExpertCG, Ashutosh2024ExpertAFEA, Panchal2024WhatTS}. 

Providing feedback on athlete performance is a challenging but impactful problem for Video-LLM models \cite{Tang2023VideoUW, wang2025internvideo, Qwen25VL}. 
This is a relatively new task and involves generating targeted and actionable guidance on the player's actions in a video. This feedback could reinforce \emph{positive movement patterns} (e.g. ``The climber has good positioning with their right foot on the hold, allowing them to turn their right hip in and reach up.''), identify \emph{areas for improvement} (e.g. ``The player's use of the bottom of the foot roll to the outside of the body is causing balance issues and slowing them down.''), and/or \emph{issue corrective guidance} (e.g. ``The player needs to bring the ball placement higher above his shoulders to prevent defense from stripping the ball and to finish more efficiently.'').  This form of text generation goes beyond action recognition or captioning of broad events occurring in a video, and requires a combination of domain-specific action, biomechanical, and quality understanding.

\input{figures/concept}

Unfortunately, collecting this high-quality sports feedback requires experts to annotate videos which is prohibitively expensive. While there have been massive efforts to collect this data, such as EgoExo4D \cite{Grauman2023EgoExo4DUS} and ExpertAF \cite{Ashutosh2024ExpertAFEA}, this data is limited in action coverage even after extensive effort. First, it is limited to only three sports: soccer, basketball, and rock climbing. Second, the basketball and soccer videos consist of drills for shooting and dribbling, rather than demonstrations covering the full action space of soccer and basketball. Since the domains (specific sports) of the annotated data are limited, studying generalization of Video-LLMs finetuned on a particular source domain to an unseen target domain is important.

We propose finetuning Video-LLMs on abundant, freely available auxiliary multimodal data from the target domain alongside annotated sports feedback from the source domain. While specific movement patterns differ across sports, fundamental skills like timing, coordination, power, and balance could transfer between domains. For instance, the hip flexion and explosive leg drive used in basketball jumping applies similarly to power generation for upward movement in bouldering. To capture domain-specific knowledge, we leverage two key sources of auxiliary data: competition videos with weakly-aligned live, narrated commentary and books or manuals that contained actionable advice and domain-specific knowledge about biomechanics and movement details. The commentary, in particular, captures \emph{detailed information about body part positioning} (e.g. ``Right hand holds the board firmly for more distance, while left hand holds it loosely.''), insights about \emph{better movements and more optimal biomechanics in that moment} (e.g. ``...I would engage my shoulder to go up...'') and more generally \emph{domain-specific terminology} like ``Fujii adjusted his right foot onto the right \emph{sloper} and then paused.''

However, leveraging such commentary data effectively remains non-trivial. Commentary often varies widely in relevance and quality, and is associated over a long window in the video, rather than the precise moment the action occurred. % and text-only data can degrade the model’s ability to follow instructions and interpret visual content. % when used naively. 
We address the challenge of using competition video commentary data through our proposed LLM-based data filtering and rewriting technique (refinement). It aims to select and summarize the most relevant segments from the freeform commentary data and improve alignment of the refined commentary to timestamps in the video (precise localization). Further, expert commentators are more descriptive, rather than prescriptive in their commentary. We show that this can bias the style of the feedback, and show how training with the text-only data reduces this bias. %We show that a two-stage finetuning approach, specifically text-only finetuning followed by the video-text finetuning, is the best way to incorporate the text data. 

We also find limitations in the evaluation of sports feedback generation. Standard reference-based text generation metrics will capture how well the generated text matches the reference text. While this may indirectly indicate feedback quality, we believe it is important to capture aspects of feedback quality for interpretability. Even metrics that capture semantic similarity may fail to account for how specific and actionable the feedback is. For instance, a generated statement such as
``The shooter is shooting the ball poorly'' is semantically aligned to the reference, ``The ball’s trajectory is flat because the release point is too late. This is because the shoulders and hips are slow to rotate, so rotate the hips faster.'' Despite being semantically aligned, the generated statement lacks rich details about the current performance and contains nothing actionable. In contrast, the annotated feedback provides specific details about the movement pattern and the impact on movement quality, and causal details about the body parts and motion that cause this failure. 

To enhance the interpretability of evaluation, we propose two LLM-based automatic evaluation metrics,  specificity and actionability, which are grounded in motor learning theory \cite{schmidt2019motor}. The feedback targeted in this work aligns with Knowledge of Performance (KP) feedback,
which focuses on movement quality rather than outcomes. KP includes descriptive feedback (explaining what occurred) and prescriptive feedback (suggesting improvements). Specificity measures descriptive content, while actionability assesses how implementable the suggested corrections are. We also validate LLM performance as automatic evaluators for both metrics with human performance. 
% with Knowledge of Performance (KP) feedback, which provides information about the movement itself, rather than the outcome. KP can be further categorized as descriptive, which explains what occurred during the movement and its quality, and prescriptive, which suggests how to correct or improve the movement. Specificity captures the degree of descriptive KP content, while actionability is associated with prescriptive KP and measures how implementable the provided corrections are.  We also validate the performance of LLMs when used as automatic evaluators for these two aspects prior to usage. %
%We see that these metrics allow for a deeper analysis about model generation behavior that was not visible in the existing metrics.

To summarize, our contributions are:
\begin{enumerate}[nolistsep,noitemsep]
    \item collecting and refining free-form competition multimodal data for sports feedback usage;
    \item improving out-of-distribution sports feedback generation performance by 106\% on BLEU-4, 36\% on METEOR, 39\% on ROUGE-L, and 25\% on BERTScore with multimodal and unimodal freely available data compared to only finetuning on out-of-distribution sports feedback data;
    \item and proposing LLM-based reference-free evaluation metrics that are specific to sports feedback generation and validating on human-annotated data.
\end{enumerate}

%% file: figures/concept.tex
\begin{figure}
\centering
  \includegraphics[width=0.95\linewidth]{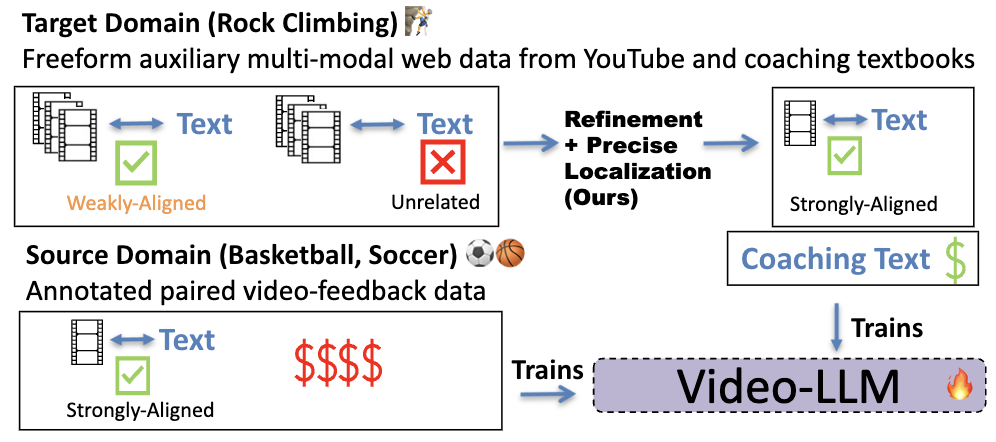}
  \caption{Our method leverages expensive, strongly-aligned annotated video-feedback pairs from a source domain (basketball, soccer) alongside abundant, freely available auxiliary data from the target domain (rock climbing). The auxiliary data includes weakly-aligned and unrelated video-text pairs from YouTube and coaching textbooks. Through refinement and precise localization, we transform weakly-aligned into strongly-aligned training data, enabling effective cross-domain transfer in data-scarce settings.}
  % \caption{Existing sports feedback generation data is limited in sports coverage because it is expensive to collect. We focus on generating feedback for an unseen target domain (rock climbing). We demonstrate that training video-LLMs on (1) existing sports feedback from a set of source domains and (2) freely available auxiliary web data (rock climbing competition videos and coaching texts) from the target domain, can reduce the domain generalization gap to the target domain significantly.}
  % \caption{Existing sports feedback generation data is limited in sports coverage because it is expensive to collect. We focus on generating feedback for an unseen target domain (rock climbing). We demonstrate that training video-LLMs on (1) existing sports feedback from a set of source domains and (2) freely available auxiliary web data (rock climbing competition videos and coaching texts) from the target domain, can reduce the domain generalization gap to the target domain significantly. We also find that traditional text generation metrics do not capture fine-grained aspects of feedback quality. To address this, we propose automatic evaluation of specificity and actionability.}
  \label{fig:teaser}
\end{figure}

%% file: sections/related_works.tex
\section{Related Works}

\textbf{Sports Commentary Generation.} 
Most professional sports competitions are accompanied by commentary which provides viewers with background context, analysis of the game, and entertain viewers. Commentary generation is a video-language task to generate commentary given video input. Prior work uses written commentary data which are minute-by-minute textual updates in matches \cite{Mkhallati2023SoccerNetCaptionDV, Rao_Wu_Liu_Wang_Xie_2024} or annotated dense captions \cite{Yu2018FineGrainedVC}. We collect data from live-speech commentary, so the commentary text is derived from noisy ASR transcripts accompanying competition videos. We find that, despite the noise, the live-speech commentary goes beyond dictating events, and contains rich quality and movement analysis information. %, and so, is not limited to the sports or games  sports/games transcribed by a particular platform and is more scalable than annotating commentary manually. 
Other prior work uses commentary generation as an additional objective \cite{Parmar2019WhatAH} or weak supervision \cite{Du2024LearningSR} to support action quality assessment. Our purpose is to \textbf{reduce reliance on labeled feedback data} for sports feedback generation by using commentary data and by extension, learning commentary generation. %Specifically, we leverage actual feedback data from some sports, and commentary from others without feedback, to enable the generalization of feedback generation to new sports without annotated feedback. 

\textbf{Sports feedback generation.} Sports feedback generation is the task of generating targeted and actionable guidance on the player's actions in a video. Commentary and feedback generation have distinct characteristics. Commentary generation typically focuses on describing events and engaging viewers. In contrast, feedback offers evaluative or corrective insights aimed at improving performance. This is a relatively recent task, first explored in \cite{Zohar2024VideoSTaRSE}, which uses chain-of-thought reasoning to analyze performance before predicting a score, though it does not assess the quality of the analysis or ensure actionable feedback. \cite{Ashutosh2024ExpertAFEA} is the first to explicitly define the task and introduces a curated feedback dataset, ExpertAF. \cite{Seino2025ExpertCG} further proposes an architecture that incorporates spatio-temporal features and models skill level. Our work is orthogonal to these efforts, and focuses on the generalization of sports feedback to unseen sports, with the goal of reducing the need for costly annotated feedback data.

\textbf{LLMs for automatic evaluation.} Prior work in sports feedback and commentary generation has explored using large language models (LLMs) for automatic evaluation. \cite{Rao_Wu_Liu_Wang_Xie_2024} uses GPT-Score \cite{fu-etal-2024-gptscore}, which rates how likely a generated text aligns with a given evaluation protocol (e.g., task specifications, aspect definitions) on a 1–10 scale. Similarly, \cite{Panchal2024WhatTS} proposes LLM-Acc, scoring from 1–5 how closely the output matches a reference text. In contrast, we use LLM-based scoring to focus specifically on \emph{unique, feedback-centric properties} such as, specificity and actionability. This offers more interpretable, aspect-level scores rather than a single holistic metric and is independent of the reference. Furthermore, we also validate these metrics on feedback data unlike \cite{fu-etal-2024-gptscore} which evaluates on general natural language understanding tasks like summarization, dialog, and machine translation.

\textbf{Learning from noisily aligned video-text data.} HowTo100M \cite{Miech2019HowTo100MLA} is a large-scale, uncurated video-language dataset from Youtube that is often used for video-language representation learning \cite{Lin2022LearningTR, Miech2019EndtoEndLO, Xu2021VideoCLIPCP, Zhou2023ProcedureAwarePF}. However, these weakly-aligned video segment-narrations introduce two key challenges: low-quality or off-topic transcriptions, and temporal misalignment between video and text. To address the former, prior work either replaces ASR entirely with captions generated by video-language models \cite{zhao2022lavila}, or refines ASR output by prompting LLMs to produce more coherent narrations \cite{shvetsova2023howtocaption}. Others apply vision-language models like CLIP \cite{Radford2021LearningTV} to filter out poorly aligned pairs \cite{Wang_Meng_Liang_Wang_Liu_Zhao_2024}. We follow a similar LLM-based filtering strategy, but specifically prompt the model to extract concise, feedback-relevant commentary from noisy ASR transcripts, rather than augment the caption with new information.

Temporal misalignment is harder to correct. Existing methods rely on strong assumptions from procedural video domains—such as consistent step order and single-step execution per video \cite{Chen2024LearningTL, Han2022TemporalAN}, which do not hold for competition footage involving open-ended gameplay or multiple attempts by different competitors (e.g., figure skating, rock climbing). Commentary often includes nuanced motion cues and terminology missed by general vision-language models \cite{Radford2021LearningTV, Xu2021VideoCLIPCP}, \cite{Rao_Wu_Liu_Wang_Xie_2024} avoids this and uses an existing soccer-specific commentary generation model for ASR narration localization. % addresses this by generating dense captions with an existing soccer-specific commentary generation model to generate dense captions (caption for each frame) and then coarsely-aligning narrations from the transcript. 
We adopt a simpler approach, meant to be applicable to any sport: prompting an LLM to refine long, noisy ASR transcripts into concise commentary, then coarsely aligning this refined text to word-level timestamps using Whisper. This approach avoids reliance on dense captioning or grounding from pre-trained video-language models which might be weaker on commentaries, and enables us to begin the use of commentary data to support feedback generation in sports without labeled feedback.

%% file: sections/dataset.tex
\section{Auxiliary Multimodal Dataset}
\label{dataset}

% In a practical setting, one may wish to generate feedback for a sport where extensive, annotated feedback is not available. To explore the feasibility of this task, we study whether feedback generation can generalize to unseen sports, using basketball and soccer as source domains and evaluating on rock climbing as the target. To aid generalization to this target domain, we collect auxiliary data that is freely available on the web. We prefer to use competition web videos, since they are abundant and frequently have experts providing rich, fine-grained commentary related to the quality of actions or include domain-specific action and object-interaction terminology (e.g. ``types of rock climbing holds''). We also explore the use of text-only data using books from the target domain. We focus on the target domain of rock climbing, as the other two possible domains, soccer and basketball have a larger domain gap between competition videos and the sports feedback dataset \cite{Ashutosh2024ExpertAFEA}. This is because the competition videos are in a team setting, whereas the feedback data is collected for individual drills. 
% In contrast, both the competition videos and the sports feedback videos in \cite{Ashutosh2024ExpertAFEA} feature climbers solving rock climbing problems, rather than doing drills. 
% While we focus on rock climbing, our data collection is general and could be applied to many sports.
In a practical setting, one may wish to generate feedback for a sport where extensive, annotated feedback is not available. To explore the feasibility of this task, we study whether feedback generation can generalize to unseen sports, using basketball and soccer as source domains and evaluating on rock climbing as the target. We collect freely-available auxiliary web data to aid generalization, focusing on competition broadcasts with expert commentary that provides rich, fine-grained feedback on action quality and includes domain-specific terminology. We also explore text-only data from target domain books. For evaluation, we are limited to ExpertAF\cite{Ashutosh2024ExpertAFEA} which contains individual demonstrations across rock climbing, soccer, and basketball with feedback annotations. We choose rock climbing as our target domain over soccer or basketball because rock climbing competition \textit{broadcasts} are more similar to the rock climbing \textit{demonstrations} in ExpertAF. In comparison, soccer/basketball \textit{broadcasts} feature \textbf{team}/game performances which have a larger domain gap with the \textbf{individual} skill \textit{demonstrations} for soccer/basketball present in ExpertAF \cite{Ashutosh2024ExpertAFEA}.
%We choose rock climbing because it has a smaller domain gap between competition broadcasts and the only dataset for sports feedback \cite{Ashutosh2024ExpertAFEA}. 
% Specifically, soccer/basketball competition videos show game play whereas the sports feedback dataset \cite{Ashutosh2024ExpertAFEA} only contains individual drills. 
While we focus on rock climbing, our data collection approach is general and applicable to many sports.

We collect 18,615 video clip-commentary pairs from rock climbing competition videos on the web and free-form text (97,989 tokens) scraped from a rock climbing self-coaching manual. In this section, we describe our method for collecting this data and ensuring the quality of this data, specifically the video clip-commentary pairs. 

\input{figures/method}

% \subsection{Data collection}

\subsection{Competition video-commentary data collection} 

Video-sharing platforms like YouTube contain broadcasts of sports competitions. These broadcasts have expert commentators who provide analysis of the athlete's performance and details like the sport-specific actions or moves. This type of data is often neglected by the video-language community in favor of instructional videos \cite{Miech2019HowTo100MLA} which typically, contain broadly applicable actions and contain procedural events (e.g. ``cut the onions'', then ``mix the onions with tomato'') to accomplish some task. In contrast, the commentary data discusses domain-specific actions in detail (e.g. ``The climber holds a sphere with both hands, with the right hand applying gentle pressure.'') which is more closely related to feedback data. 

To emphasize retrieving competition broadcasts over vlogging, entertainment, or short rock climbing content where the narration might not carry feedback-relevant information, we select five channels that often return from the keyword ``rock climbing competition''. This results in 2.3K videos, and to avoid downloading videos unrelated to competition videos, such as shorts, we filter out videos shorter than 20 minutes, leaving 1,440 videos. We find this is a reasonable assumption upon manual inspection. 

\textbf{Creating video clip-commentary pairs.} Each competition video is accompanied by ASR transcripts of expert commentary. We access these ASR transcripts, which are speech-to-text transcriptions, from the Youtube Data API. The ASR transcript contains text chunks with narration start and end timestamps. Like prior work in using large-scale instructional videos \cite{Miech2019HowTo100MLA}, the video segment associated with the start and end timestamps is weakly-aligned to the text chunk. We say they are weakly-aligned because the commentators usually discuss the video context occurring at that narration timestamp. This can raise problems because: (1) the commentator may discuss either background context or offer little groundable information (e.g. ``Ooh nice nice...''); (2) the narration may be in anticipation of/reaction to an action; or (3) the associated video segment may be too long (see large range in ASR timestamps in Fig. \ref{fig:method}).
% since precise timestamps marking the video segment are lost in the ASR transcripts from the Youtube Data API. 
We address the first and third of these problems in our approach, see Fig. \ref{fig:method} (LLM Refinement and Precise Localization, respectively). \begin{comment}Since the video segment span on average ~13 seconds, we divide the video segment into smaller 4-second segments to allow for dense sampling without significantly increasing computational complexity. This is also the same video segment length as the sports feedback videos used in \cite{Ashutosh2024ExpertAFEA}.
We extract the 4-second video segments using a sliding window with a 1-second stride, and weakly pair each segment with the corresponding commentary.
\end{comment}

\textbf{Refinement. }
To address the first problem, we prompt an LLM to (1) classify if the narration only contains irrelevant content and discard the narration and if not, (2) summarize the narration to only contain action quality-relevant information. The LLM filters out 80\% of narrations, leaving 75K narrations. We only worked with about 18K narrations due to YouTube restrictions. We use the Phi-4 14B large language model \cite{Abdin2024Phi4TR} for this refinement procedure and provide the prompt in supp.
After we apply LLM refinement to filter unintelligible/irrelevant content (80\% of transcript) and improve clarity, we use Whisper to obtain precise word-level timestamps for the selected narrations.

\subsection{Text-only data collection} LLMs acquire some commonsense reasoning and knowledge from simply being trained to predict the next text token \cite{brown2020language}. Motivated by this as a mechanism to acquire rock climbing knowledge in an inexpensive way, we collect high-quality text data from a popular rock climbing coaching manual: ``Self-Coached Climber: The Guide to Movement, Training, Performance'' \cite{hague2006self} for text-only tuning.\footnote{We have bought a physical copy, but use a PDF copy sourced from LibGen (\url{https://libgen.is}) for easy text-extraction.} 
% While we focus on rock climbing for our experiments, our data sources for collection are scalable. 

% We use InternVideo2.5 \cite{wang2025internvideo}, a video-LLM for our experiments.

%% file: figures/method.tex
\begin{figure*}
    \centering
    \includegraphics[width=0.9\linewidth]{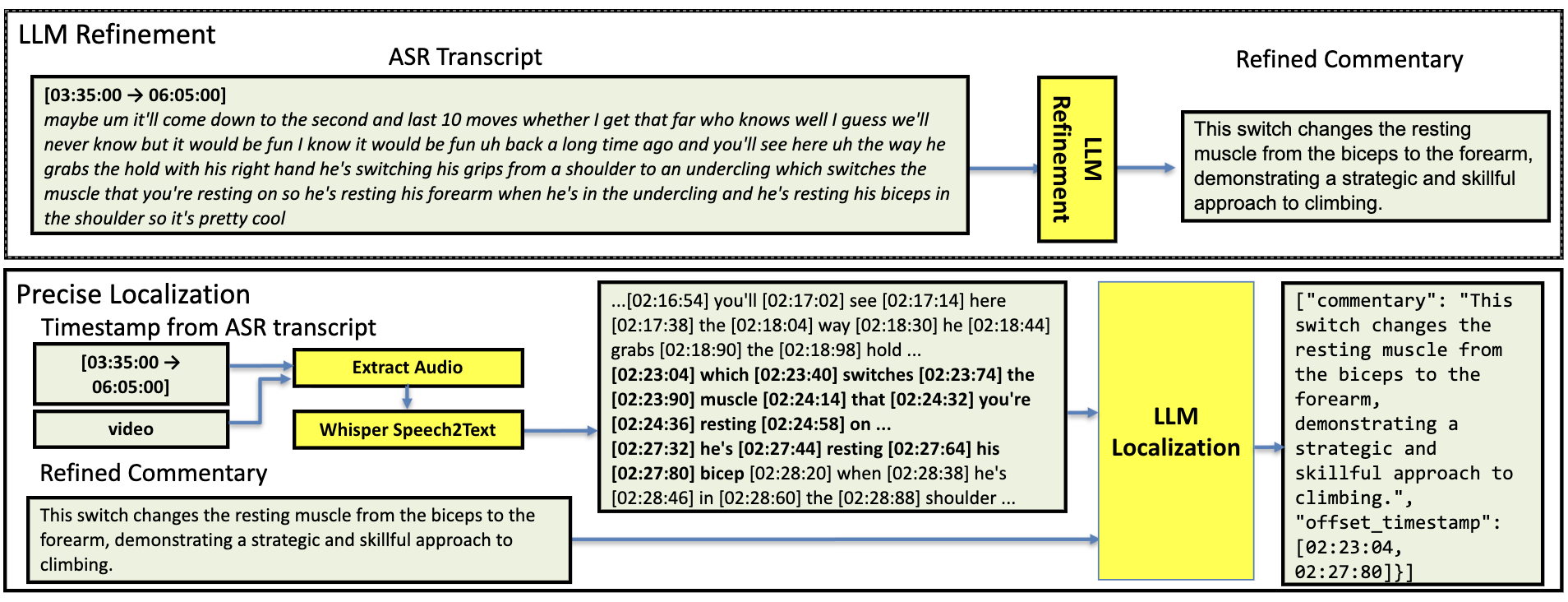}
    \caption{Two-stage process to improve the quality and temporal localization of commentary. Top: The original ASR verbose text transcript is passed to an LLM prompted to summarize the commentary concisely and extract only action-relevant or action quality-relevant information, or skip if no such information is present. Bottom: Our precise localization technique is applied to refined commentary. For each refined commentary, the corresponding ASR timestamps are used to extract audio, then passed to Whisper to obtain word-level timestamps \textbf{relative to the ASR segment start timestamp}. Finally, the refined commentary and the word-level timestamps are passed to another LLM prompted to localize where the refined commentary is narrated relative to the ASR segment start timestamp. The output is a list of commentary segments (as each refined commentary may contain multiple parts) with precise offset timestamps.}
    \label{fig:method}
\end{figure*}

%% file: sections/method_rewrite.tex
\section{Method}
\label{method}

We first discuss preliminaries of text generation conditioned on video input and formally state the feedback generation task. We then introduce our proposed precise localization method for computing a more accurate timestamp for the refined commentary compared to the timestamp provided by the ASR transcript. Finally, we explain our approach to unifying supervision from auxiliary sources in the target domain with annotated feedback from the source domain.

\subsection{Preliminaries: Video-LLM}

Before discussing finetuning, we first describe the video-LLM architecture used in our experiments \cite{wang2025internvideo}. The model follows a standard architecture with a vision encoder that transforms each video frame into a sequence of patch embeddings. Tokens from frames within a clip (4-frames) are then grouped and merged based on their pairwise embedding similarity. Then, a multilayer perceptron (MLP) projects these merged visual embeddings into the language model's input space. The resulting sequence is fed into an LLM, which autoregressively predicts the next token given the preceding context.

Formally, we define the feedback generation task as follows. Given a dataset \( \mathcal{D} = \{(v_i, t_i)\} \) consisting of 4-second video segments \( v_i \) paired with corresponding feedback text \( t_i \), the goal is to train a video-conditioned language model that generates feedback \( \hat{t}_i \). The video-LLM is finetuned using an autoregressive next-token prediction (NTP) objective, where the model learns to predict each token in the output sequence conditioned on both the video input and previously generated tokens.

\subsection{Precise localization of refined commentary}

ASR text chunks are typically paired with an entire video segment, but the refined commentary may only refer to a small sub-segment as shown in Fig. \ref{fig:method}. This misalignment arises because ASR tools produce timestamps for larger text chunks rather than individual words, and prior work \cite{zellersluhessel2021merlot} has shown that these chunks often contain redundant or repeated content. Merging such overlapping transcripts helps eliminate repetition, but in doing so, further broadens the temporal span of the associated video segment—reducing precision of the ASR timestamps.

To address this, we propose a two-step alignment strategy to localize the refined commentary more precisely. First, on the refined video-commentary subset, we re-compute the transcript using Whisper-Large-v3 \cite{Radford2022RobustSR} which provides word-level timestamps rather than text-chunk level timestamps. Then, we align the cleaned feedback-like summary (produced during refinement) to its most likely location in the transcript using a large language model \cite{Abdin2024Phi4TR}. 
%This text-only alignment process is inspired by \cite{ActionAtlas}, which similarly combines Whisper-based re-segmentation with LLM-based alignment for fine-grained localization. Our work differs in it's application to detailed narrations, compared to this work solely focusing on curating a fine-grained action benchmark.

\subsection{Unified Supervision Training Objective}

We treat ``Basketball'' and ``Soccer'' as source domains and ``Rock Climbing'' as the target domain. The video-sports feedback paired training data \( (v_i, t_i) \) comes from the source domains, while weakly aligned video segments \( \{(v_j, \tilde{t}_j)\} \) are collected from web-scraped target domain competition footage with co-occurring commentary. We also include text-only supervision \( \{t_k\} \) sourced from target domain-specific coaching manuals. Since video inputs are padded by \texttt{<image>} tokens in the input sequence, the LLM is aware of when visual tokens are included, hence text-only sequences are not a problem and all forms of input can be trained with the same NTP objective.

To recap the pipeline thus far, the following data sources are used for supervision during finetuning:

\begin{itemize}
    \item \textbf{Source Domain Sports Feedback:} Video clips from basketball and soccer drills paired with expert coaching comments, collected from the ExpertAF dataset \cite{Ashutosh2024ExpertAFEA} (sourced from Ego-Exo4D \cite{Grauman2023EgoExo4DUS}). These are high-quality (video, text) pairs focused on performance correction.

    \item \textbf{Target Domain Refined Competition Clip-Commentary Supervision:} Rock climbing competition clips \( v_j \) paired with ASR transcripts \( t_j \), cleaned and rewritten into concise action-relevant summaries \(  \tilde{t}_j \) using an LLM (Phi-4 14B \cite{Abdin2024Phi4TR}). The summaries are temporally re-aligned using Whisper-Large-v3 to produce aligned (video, feedback) pairs in the target domain.

    \item \textbf{Text-Only Coaching Knowledge:} Paragraphs \( t_k \) from the rock climbing coaching manual \cite{hague2006self}, used to expose the model to domain-relevant concepts and terminology.
\end{itemize}

Training proceeds are as follows. All supervision is cast into a unified autoregressive format. Each input is a token sequence \( x \), which may contain $n$ tokens: visual tokens followed by text tokens, or only text tokens. The model is trained to predict the next token \( \hat{y_i} \) at each step. The actual next token \( x_{i+1} \) is used as ground-truth \( y_{i} \). Note, $y_i$ and $\hat{y_i}$ are distributions over the token vocabulary:

\begin{equation}
    \mathcal{L}_{\text{NTP}} =  \frac{1}{n-1} \sum_{i=1}^{n-1} CrossEntropyLoss(\hat{y_i}, y_i) 
\end{equation}

This formulation enables the model to learn generation conditioned on either multimodal (video-text) or text-only sequences. By unifying supervision from explicit (source domain) and weakly aligned (target domain) video-text examples, along with standalone coaching text (target domain), the model can acquire both domain-specific knowledge and capture the structure and style of feedback. To our knowledge, this is the first work to collect rock climbing-specific commentary data and explore the use of auxiliary data for sports feedback generalization.

%% file: sections/experiments.tex
\section{Experiments}

We present our experiment details, then our proposed evaluation metrics, and validate their quality. In our experiments, we show how our proposed method shows improvement in generalizing sports feedback generation and ablations. Finally, we show results on our proposed evaluation metrics.

\subsection{Experimental setup}

\textbf{Implementation Details.} We use low-rank adaptation (LoRa) \cite{hu2021lora} with rank 8 and LoRa dropout as 0.4 for efficient finetuning of the large language model in the vid-LLM, finetune the vision-language adapter multilayer perceptron, and keep the vision encoder frozen. We use a learning rate of $1e-4$ and the AdamW \cite{kingma2014adam} optimizer with 50 warmup steps and weight decay of $1e-2$.  We use InternVideo2.5 \cite{wang2025internvideo} with 4-bit quantization. We train with a batch size of 16 %, decomposed as a batch size of 2 with 8 gradient accumulation steps 
for 3 epochs on a single NVIDIA L40 GPU. 

\textbf{Datasets.}
We use the auxiliary multimodal dataset curated in Sec. \ref{dataset} as a source of commentary and textbook instructions. As a source of feedback data, we use the ExpertAF \cite{Ashutosh2024ExpertAFEA} version of EgoExo4D \cite{Grauman2023EgoExo4DUS}.

% \begin{table}[h]
%     \centering
%     \small
%     \begin{tabular}{p{0.2\linewidth} | p{0.7\linewidth}} \hline
%         \textbf{Original Expert Commentary} & ``The blue hold with the edge she’s matching with both hands is positive... it’s wide and has a groove enabling the climber to use her hands like hooks. While she spends some time allocating her feet... she is flexing her arms... activating all the muscles so she’s spending a lot of energy rather than extending her arms and laying low in rest position... This tension or flex that she’s doing can compensate her arms'' \\ \hline
%          \textbf{LLM-Generated Summary (Feedback)} & ``The climber is wasting energy by flexing her arms instead of extending them and resting while allocating her feet.'' \\
%     \end{tabular}
%     \caption{Example of original expert commentary in \cite{Grauman2023EgoExo4DUS} and the summarized feedback counterpart in Expert AF \cite{Ashutosh2024ExpertAFEA}.}
%     \label{tab:commentary_summarization}
% \end{table}

EgoExo4D \cite{Grauman2023EgoExo4DUS} contains video demonstrations and expert commentaries with feedback corresponding to specific timestamps in the video. ExpertAF \cite{Ashutosh2024ExpertAFEA} creates a physical skills subset (3 out of 7 total scenarios) and summarizes the expert commentaries into explicit feedback. %Specifically, this work uses Llama3.1 70B \cite{Dubey2024TheL3} to summarize the verbose, freeform expert commentaries 
into concise summaries.% as shown in Tab. \ref{tab:commentary_summarization}. 
Note, these commentaries are explicitly collected for the purpose of critiquing athlete performance by paid experts, whereas our collected auxiliary commentary data is crawled from the web and is free.
%(e.g. ``The climber is wasting energy by flexing her arms instead of extending them and resting while allocating her feet.''). 
% There are 10,299 rock climbing, 5,647 soccer, and 16,939 basketball feedback instances associated with a corresponding timestamp in the video. 
Following ExpertAF \cite{Ashutosh2024ExpertAFEA}, we use the preceding 4 seconds before the timestamp as video input and uniformly sample 16 frames. Empirically, we found that 16 frames performed better than 128 frames; we suspect this is because rock climbing (our target sport) is slower paced than rapid actions like a basketball shot, hence might not benefit much with denser sampling. 
% EgoExo4D/ExpertAF \cite{Grauman2023EgoExo4DUS, Ashutosh2024ExpertAFEA} provide 8 exocentric and 1 egocentric camera views per demonstration. 
We also use only the best exocentric view (based on provided metadata) and discard all other views for simplicity. Pose data is also excluded from our experiments. 

We train on 18,049 expert feedback samples from basketball and soccer, and 14,855 refined rock climbing commentaries. We validate on 4,537 basketball/soccer feedbacks and 3,760 rock climbing climbing commentaries, and evaluate on 2,806 held-out rock climbing feedbacks.

\subsection{Evaluation metrics}
\label{sec:eval_metrics}

\subsubsection{Problems with existing metrics}

Text generation is typically evaluated against a reference text written by a human (``ground truth''). Standard metrics, such as, BLEU-4 \cite{papineni2002bleu}, METEOR \cite{banerjee2005meteor}, and ROUGE-L \cite{lin2004rouge} capture the lexical similarity between the generated text and the reference text. METEOR \cite{banerjee2005meteor} allows for greater flexibility by accounting for fuzzy matches (synonyms and stems) between the generated and reference text. To evaluate semantic similarity, rather than text fluency, prior work has used BERTScore \cite{zhang2019bertscore}.
Existing metrics often fail to capture the qualities that make feedback effective. For example, lexical overlap-based metrics like BLEU and ROUGE are sensitive to exact forms and may assign low scores to feedback that is semantically accurate but phrased differently. Consider the reference, ``Your follow-through is too short, which limits power,'' and the prediction, ``Extend your follow-through more to generate power.'' Although both convey the same idea and the prediction is explicitly actionable, the prediction may receive a low score due to limited word overlap. Similarly, metrics that aim to capture semantic similarity, such as BERTScore, may not reflect the \textbf{usefulness} of the feedback. For instance, given the reference ``Try planting your foot earlier to stay balanced,'', the prediction ``Your balance could be improved'' would have high semantic similarity despite being vague and lacking concrete guidance. These examples illustrate that how existing metrics miss aspects of feedback that determine quality.

\subsubsection{Our proposed metrics}
\label{proposed_metrics}
We propose two LLM-based automatic evaluation metrics that capture important aspects of feedback quality. To motivate this, we consider the purpose of feedback drawing from motor learning theory \cite{schmidt2019motor}. Feedback has two purposes. First, it should deepen the learner's understanding of how their performance (movement patterns, quality indicators, etc) connects to outcomes, known as \textit{descriptive} knowledge of performance \cite{schmidt2019motor}. Second, if the performance requires improvement, the feedback should be actionable by the learner, known as \textit{prescriptive} knowledge of performance \cite{schmidt2019motor}. We map these types of feedback into two aspects of feedback quality as evaluation metrics and provide definitions below.
%after multiple rounds of calibration with annotators below. 

\begin{definition}[Specificity]
Specificity describes how precisely feedback communicates what is happening in the learner's movement at the current moment. Higher specificity includes detailed information about the movement (e.g., positioning, timing) and links it to quality indicators like smoothness, control, or efficiency. The most specific feedback also explains when or why the issue occurs. We use a 4-point scale where:
\begin{itemize}
  \item 1 = Not specific (e.g., vague or generic statements)
  \item 2 = Some movement or quality info, but not both
  \item 3 = Connects movement details to quality indicators
  \item 4 = Adds elaboration (e.g., cause-effect or timing cues)
\end{itemize}
\end{definition}

\begin{definition}[Actionability]
Actionability refers to if a learner can directly apply the feedback to make a change. More actionable feedback gives clear corrective directions (e.g., what to move, how to adjust). We use a 3-point scale:
\begin{itemize}
  \item Skipped = Only positive reinforcement (e.g. ``The climber demonstrates good balance'')
  \item 1 = Not actionable (vague or no useful guidance)
  \item 2 = Suggests change but lacks clear instructions
  \item 3 = Gives specific, actionable correction
\end{itemize}
% Positive reinforcement-only feedback is excluded from scoring.
\end{definition}

We train annotators using feedback examples at varying levels of specificity and actionability, full table provided in supp. For instance, level 3 specificity might state, ``The climber hesitates and takes a shorter step when reaching for the higher hold,'' while level 4 adds the consequence: ``...which limits the momentum needed to successfully grab it.'' In terms of actionability, the level 2 feedback of ``The ball’s trajectory is too flat'' identifies the issue, but level 3 offers a fix: ``Release the ball slightly earlier and follow through higher to create a better arc.''

\subsubsection{Annotator Agreement}
Two of the authors independently annotated 30 reference samples for specificity and actionability. To avoid issues of bias, the annotators do this after extensive examples and the detailed descriptions explained in Sec. \ref{proposed_metrics}.
%To assess agreement, we compute weighted Cohen’s Kappa: 0.21 for specificity and 0.91 for actionability, indicating moderate and near-perfect agreement respectively.
When computing agreement, we exclude cases where only corrective guidance is provided (specificity inapplicable) or where feedback was solely positive reinforcement (actionability inapplicable). Thus we retained 24 samples for specificity and 14 for actionability. On these sets, the weighted Cohen’s Kappa, which factors in relative differences, was 0.28 for specificity and  0.60 for actionability.

Closer inspection reveals that most disagreements were minor: for specificity, 8 out of 11 disagreements differed by only one rating point; for actionability, all 3 disagreements were also off by just one point.

\subsubsection{Verifying LLM scoring with human annotations}
% We average the ratings between the two annotators and exclude samples where specificity and actionability are not applicable. %This leaves 24 samples for specificity and 14 samples for actionability. 
\begin{table}[h]
\small
    \centering
    \begin{tabular}{c|c|c}
        \hline
        \textbf{Model}&\textbf{ Spec. Acc.  (\%)} &  \textbf{Action. Acc. (\%)} \\

        \hline
        GPT-4o & 70.8 & 85.7 \\
        Phi-4 14B & 37.5 & 64.2 \\
    \end{tabular}
    \caption{Verifying LLMs' ability to score specificity, actionability.}
    \label{tab:evaluation_ability_llm}
\end{table}

We compare two models: GPT-4o-mini (closed source) \cite{openai2024gpt4o} and Phi-4 14B (open source) \cite{Abdin2024Phi4TR}, as shown in Table~\ref{tab:evaluation_ability_llm}. We average ratings between annotators to serve as ground truth and we count an LLM score prediction as correct if it is within 0.5 of the ground truth rating. The prompts used here are based on the definitions provided earlier, and the full prompts are included in supp. 
GPT-4o-mini demonstrates strong performance, achieving 70.8\% accuracy on specificity and 85.7\% accuracy on actionability. However, Phi-4 has substantially lower performance, especially in determining specificity. These results suggest GPT-4o-mini is capable of capturing both descriptive and corrective dimensions of feedback. We use GPT-4o-mini over Phi-4 for scoring specificity and actionability based on these results. 

Additionally, we evaluate whether the model can skip scoring when the sample is out of scope (e.g., feedback is only positive reinforcement). GPT-4o successfully skips 100\% of such samples in the actionability task.

\input{sections/experiment_results_rearrangement}

% \begin{comment}
% \subsection{Experimental Results}

% \textbf{Does using auxiliary multimodal data help with out-of-distribution performance?}

% \input{tables/ood_table}

% \textbf{What type of data is more important to sports feedback generalization?}

% \textbf{Which parts of the model should be tuned?}

% \textbf{What is effect of filtering data}

% \textbf{What happens when the data is scaled?}

% \begin{table}[]
%     \centering
%     \begin{tabular}{c|c}
%          &  \\
%          & 
%     \end{tabular}
%     \caption{Training data ablation}
%     \label{tab:ablations}
% \end{table}

% % \begin{table}[]
% %     \centering
% %     \begin{tabular}{c|c}
% %          &  \\
% %          & 
% %     \end{tabular}
% %     \caption{Joint vs sequential training ablation}
% %     \label{tab:ood_perf}
% % \end{table}

% \begin{figure}
%     \centering
%     \includegraphics[width=0.5\linewidth]{qual_ex.png}
%     \caption{Qualitative examples}
%     \label{fig:qual_examples}
% \end{figure}
% \end{comment}

%% file: sections/experiment_results_rearrangement.tex
\subsection{Experimental Results}

We validate our method of using freely available, auxiliary target-domain data to improve sports feedback generation. We conduct ablations to evaluate how much each type of data contributes to the final performance. We leave additional ablations such as the order of incorporating the auxiliary data and the refinement/precise localization strategies in supp; 
these show the benefit of first training on the book data, then all video-text data, over training on book and video-text jointly. 
We also show results on our new evaluation metrics and demonstrate complementary results to existing metrics.

\input{tables/ood_lexical_sim_table}

\textbf{Does using auxiliary multimodal data help with out-of-distribution sports feedback generation?} We report the standard text evaluation metrics of BLEU-4, METEOR, ROUGE-L, and BERTScore in Tab. \ref{tab:ood_results_lexical_sim}. Our method is trained with our auxiliary commentary and book data, and basketball/soccer OOD feedback data; we also report results on a baseline of only training on basketball/soccer OOD feedback data (OOD Fd.) and an upper bound of trained on rock climbing/basketball/soccer feedback data (ID fd.). We use a temperature of 0.7 and use nucleus sampling for all our experiments, with $prob\_p=0.9$ to generate feedback. Since sampling is non-deterministic, we sample three responses and report the mean and standard deviation of the metrics for METEOR and ROUGE-L. Overall, the standard deviation is fairly low in the reported results and BERTScore and BLEU-4 have a std. deviation between 0-0.02 at that precision, so we don't include std. deviation in the table. This indicates that sampling doesn't affect the performance too much. 
We see that our method trained with auxiliary multimodal and the out-of-domain feedback data performs better than the model trained with only the out-of-distribution sports feedback data and the zero-shot model. Interestingly, the zero-shot model performs better than the out-of-distribution model, indicating there is a knowledge loss in feedback generation after finetuning. We observe this in the generated feedback, whereas the OOD model generates feedback unrelated to the target domain, instead from the source domain (e.g. generated feedback is ``The shooter's hand positioning is good, but the ball is too far behind the hand, causing the hand to open and the ball to hit the palm, affecting the shot's direction.'' and reference feedback is ``The climber is using a technique to remove excess chalk from their hands.''). These results suggest that incorporating auxiliary multimodal data helps improve out-of-distribution feedback generation, mitigating domain-shift and knowledge loss to an extent.

\input{tables/data_source_ablation_no_bleu}

\textbf{How does each auxiliary data source contribute to sports feedback generalization?} %We apply nucleus sampling with the same hyperparameter values as described earlier and sample three responses. 
In Tab. \ref{tab:data_source_ablation} we see that finetuning on all auxiliary data has the best performance. While finetuning on text alone has marginal improvement over no finetuning, we observe a complementary effect when the text data is combined with the commentary and OOD feedback. Combining commentary with OOD feedback sees the largest improvement relative to the zero-shot performance. 
% This aligns with the knowledge loss described earlier; 
Training with the commentary data could possibly mitigate the knowledge loss in visual grounding for the target sports.

% \textbf{How is generalization affected by commentary data scale?}

\input{figures/actionability_specificity_table}
\textbf{What is the performance on actionability and specificity?}
In Fig. \ref{fig:actionability_specificity_results} we report actionability and specificity scores using three high-quality LLMs (GPT-4o-mini, Gemini 2.5 flash-lite, DeepSeek Chat).  %All outputs are generated using nucleus sampling. 
Our full method, trained with both auxiliary sources (text and video-text) and out-of-domain feedback, consistently improves over the zero-shot baseline on both metrics.  % and even outperforms the ID model on actionability.

% The OOD model performs comparably to the ID model and even scores higher in terms of actionability. This can be partially attributed to the reference-free nature of the metrics, which evaluate each piece of feedback in isolation, without assessing its relevance to the specific demonstration. As a result, domain-irrelevant outputs from the OOD model are not penalized. We also hypothesize there are differences in the training data distributions, particularly between the basketball and soccer feedback data (OOD) and the rock climbing data added to the ID training set (which includes all three sports). To investigate, we compute actionability scores on 1,000 reference feedback examples and find that ``Basketball'' and ``Soccer'' examples have a higher mean actionability score of 1.90, compared to 1.65 for ``Rock Climbing''. This suggests that the ID model may inherit a bias toward less actionable feedback for rock climbing videos, as a result of the lower-actionability feedback associated with rock climbing in the training data. 
% Unlike actionability, both OOD and ID models have similar performance on specificity. This is reasonable because this is a reference-free metric. 

Adding auxiliary text data (see last bar) yields notable gains in actionability compared to only adding the auxiliary commentary data, possibly because the book content includes actionable information with clear cause-effect descriptions (e.g., %``Losing control of the center or making poor decisions about where to place it can lead to a serious loss of efficiency. \textbf{
``A fall becomes more likely because your center is actually pulling you off the rock.'', 
%``In this situation, the base is small, and the rising center naturally destabilizes the body. 
``If you have an adequate foothold available in the right position, %\textbf{
a drop knee with the right leg will significantly broaden the base so that the center remains within the base for the entire move.''). Note the large gains in actionability due to the use of text, compared to the previously shown metrics. This shows the \emph{complementarity of our proposed metrics} with respect to prior metrics. Specificity improves minimally with the addition of text, possibly due to its greater dependence on the visual context.
% While our models outperform the ID baseline in terms of actionability, a domain gap remains for specificity. 
The nuance in these findings further support that actionability and specificity are complementary to traditional lexical and semantic similarity metrics, offering a more complete view of feedback quality and style.

%% file: tables/ood_lexical_sim_table.tex
\begin{table}[h]
\small
    \centering
    \begin{tabular}{l|cccc}
    \hline
     & \textbf{BLEU-4} & \textbf{METEOR} & \textbf{ROUGE-L} & \textbf{BERT}\\
    \hline
    \rowcolor{gray!20} ID Fd. & 3.31  & 20.94 $\pm$ 0.14 & 25.91 $\pm$ 0.13 &  37.3\\
    Zero-Shot & 1.75  & 15.08 $\pm$ 0.12 & 19.78 $\pm$ 0.04 & 30.3\\
    OOD Fd. & 1.30 & 11.45 $\pm$ 0.12 & 17.30 $\pm$ 0.11 & 25.4\\
    Ours & \textbf{2.68 } & \textbf{15.59 $\pm$ 0.14} & \textbf{24.01 $\pm$ 0.05} & \textbf{37.9}\\
\end{tabular}
    \caption{Out-of-distribution feedback generation evaluation. ID represents an upper bound using in-domain feedback data. Bold represents best result in column. Fd=Feedback. Ours is a combination of training on OOD feedback and our auxiliary sources.}
    \label{tab:ood_results_lexical_sim}
\end{table}

%% file: tables/data_source_ablation_no_bleu.tex
\begin{table}[h]
\small
    \centering
    \begin{tabular}{l|ccc}
    \hline
    \textbf{Data Type} & \textbf{METEOR} & \textbf{ROUGE-L} & \textbf{BERT} \\
    \hline
    Zero-Shot & 15.08 $\pm$ 0.12 & 19.78 $\pm$ 0.04 & 30.3 \\
    Text  & 15.22 $\pm$ 0.06 & 19.74 $\pm$ 0.04 & 30.4 \\
    Commentary, Fd. & 15.38 $\pm$ 0.10 & 23.39 $\pm$ 0.06 & 37.0 \\
    Text, Com., Fd.  & \textbf{15.59} $\pm$ 0.14 & \textbf{24.01} $\pm$ 0.05 & \textbf{37.9} \\
    \hline
    \end{tabular}
    \caption{Data source ablation. Adding commentary data boosts performance over zero-shot and text-only fine-tuning. Training with all data types performs best across all metrics.Com=Commentary, Fd=Feedback.}
    \label{tab:data_source_ablation}
\end{table}

%% file: figures/actionability_specificity_table.tex
\begin{figure}
    \centering
    \includegraphics[width=0.8\linewidth]{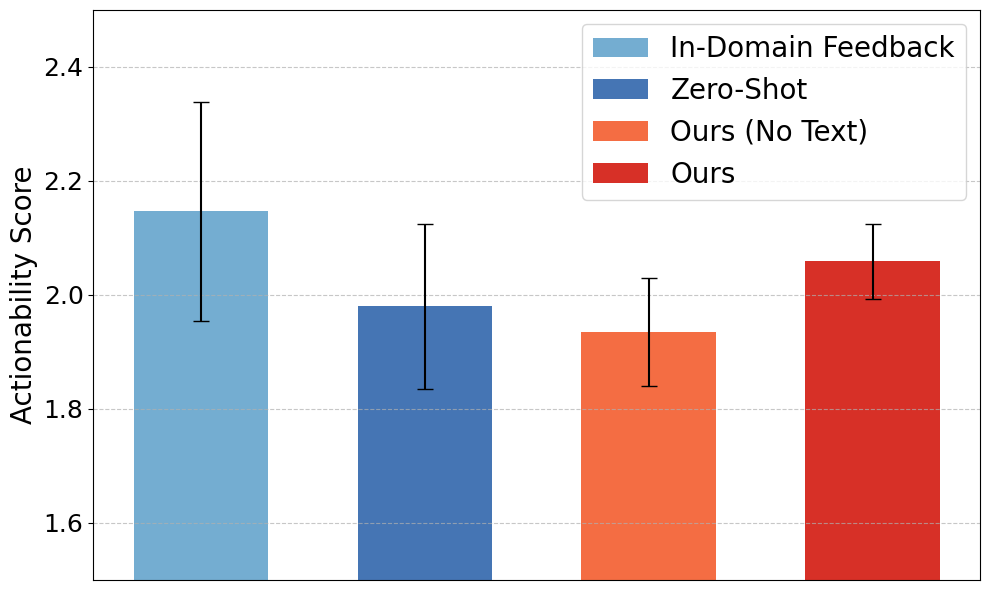}
    \includegraphics[width=0.8\linewidth]{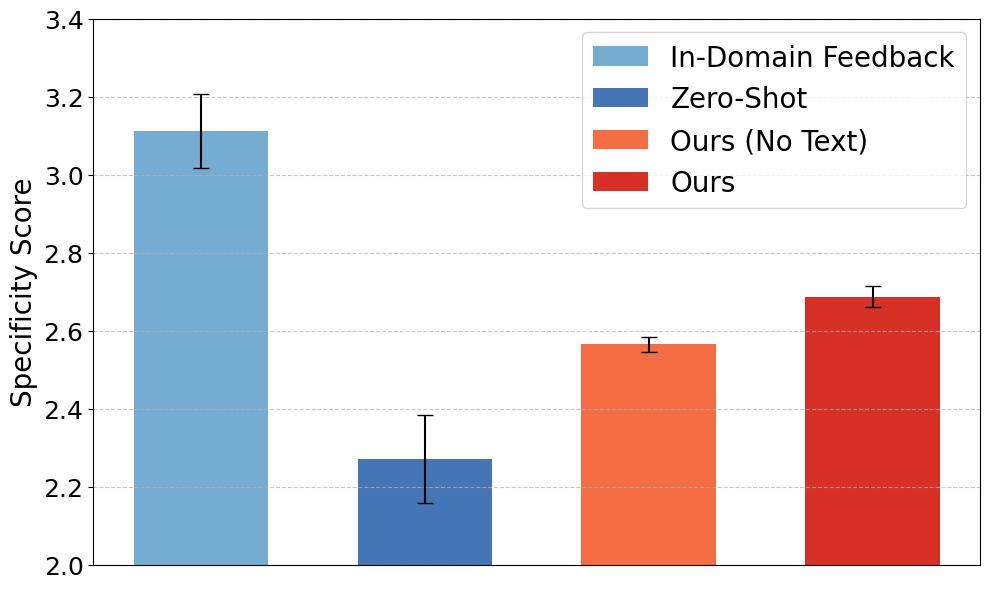}

    \caption{LLM-based evaluation of actionability (top) and specificity (bottom) as introduced in Sec. \ref{sec:eval_metrics}. The lines indicate max/min over using GPT-4o, Gemini 2.5, and DeepSeek Chat.}
    \label{fig:actionability_specificity_results}
\end{figure}

%% file: sections/conclusion.tex
\section{Conclusion}% and Future Work}
We show how to leverage freely-available competition commentary and freeform text data for generalizing sports feedback generation models to unseen sports. 
To evaluate feedback quality beyond standard lexical and semantic similarity metrics, we introduced two novel, feedback-centered metrics, actionability and specificity, grounded in motor learning theory. Our evaluation shows that models trained with auxiliary text data achieve higher actionability on the target domain compared to only using the auxiliary commentary data, while specificity remains more dependent on visual context. These metrics complement traditional scores and provide a richer picture of feedback quality. 
While we focus on rock climbing, future work could explore how to incorporate team-sport competitions, such as basketball and soccer, as auxiliary supervision. 
We acknowledge that using LLMs to automatically compute specificity and actionability has limitations, but find limited gender or length bias in our experiments conducted in supp.

\textbf{Acknowledgement. }This work was partially supported by National Science Foundation grant 2046853.

%Furthermore, there is a domain gap between the competition videos and sports feedback videos due to skill-level. The competition videos have world class athletes, while the sports feedback videos are amateurs of varying skill levels. It would be worth exploring utilizing tutorial videos which may contain visually groundable feedback that targets beginner and intermediate skill levels better than competition videos.

%Motor learning theory suggests that the effectiveness of descriptive (more specificity) knowledge of performance and prescriptive (more actionability) knowledge of performance vary based on skill level \cite{schmidt2019motor}. This could open an impactful direction in adapting feedback generation to account for this preference and adapting the LLM-based scoring approach to training (e.g. reward models).

%% file: sections/supplemental.tex
\section{Supplemental Section for Generalizing Sports Feedback Generation by Watching Competitions and Reading Books: A Rock Climbing Case Study}

\subsection{Commentary and video timing}

\begin{figure}[h]
\centering
\begin{minipage}{0.45\linewidth}
\includegraphics[width=\linewidth]{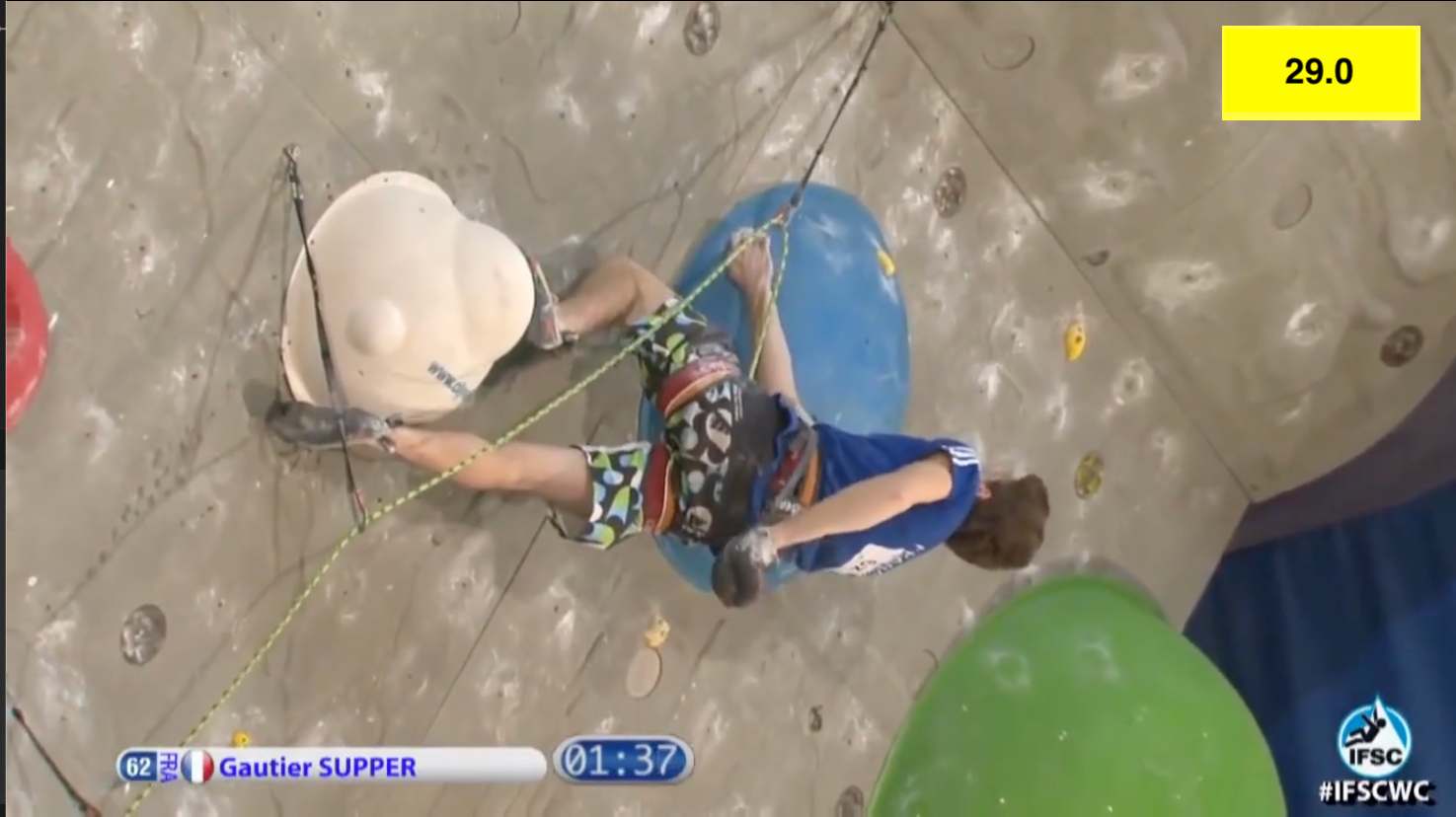}
\caption*{Looks up to anticipate the next moves. 
\textbf{Timestamp (s): [28.0, 30.26]}}
\end{minipage}
\hfill
\begin{minipage}{0.45\linewidth}
\includegraphics[width=\linewidth]{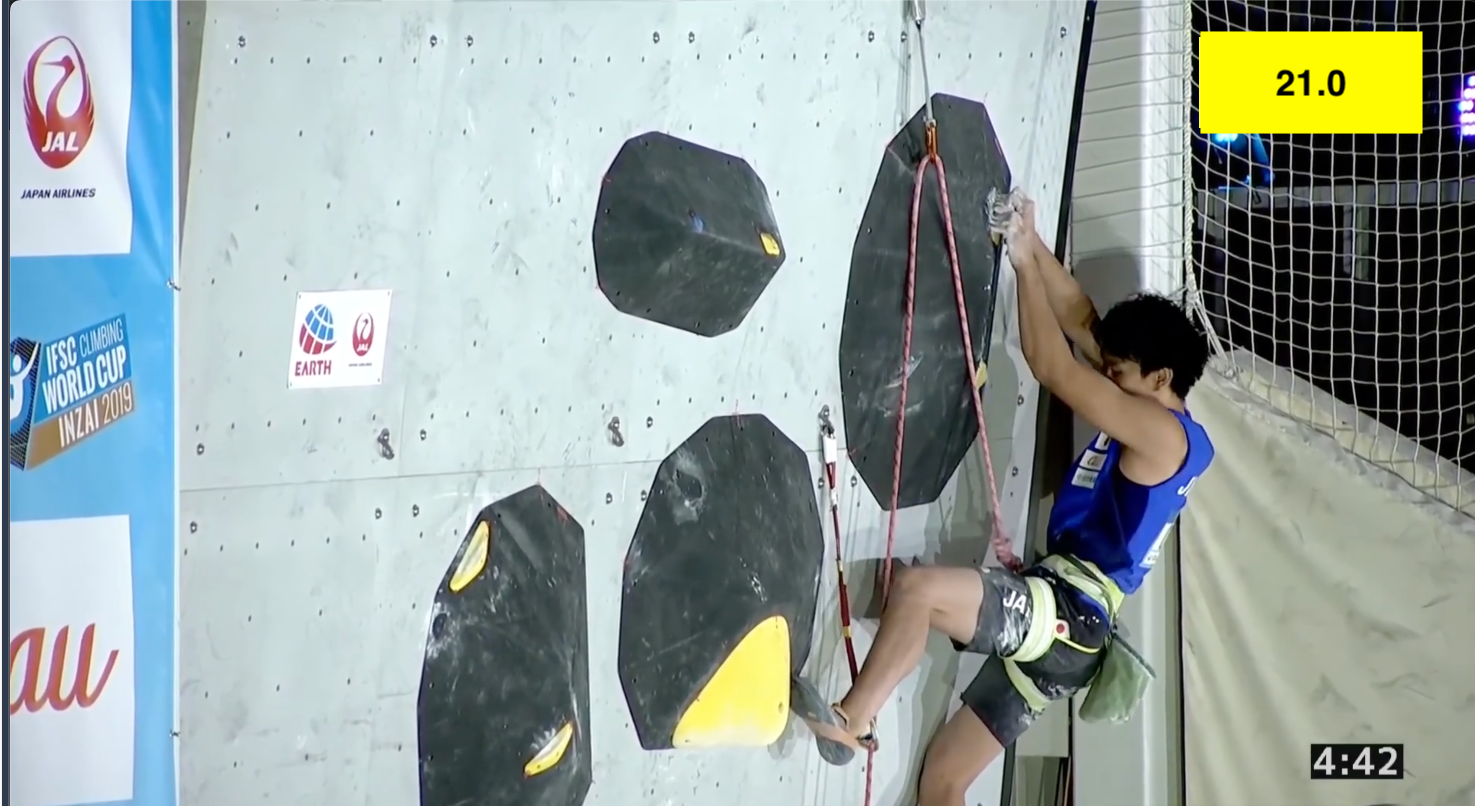}
\caption*{Climber is performing a crimp, making it appear effective. 
\textbf{Timestamp (s): [19.66, 23.66]}}
\end{minipage}
\caption{Examples of key-frames associated with refined commentary and timestamp window from precise localization step.}
\label{fig:window_ablation_examples}
\end{figure}

Our proposed precise localization method aims to pinpoint when the refined commentary occurs based on word-level timestamps. This solves temporal misalignment caused by the coarseness of the original ASR timestamps, but there still can exist temporal misalignment between the narration and the corresponding video segment. However, this is less problematic for rock climbing since climbers often remain in position for extended periods, and commentary typically addresses actions that persist over time. We demonstrate this property with examples in Fig.~\ref{fig:window_ablation_examples}. For sports beyond rock climbing that have short, fast-moving actions or events, finer temporal localization methods may be required (e.g., predicting the offset from the localized narration timestamp).

We also show an ablation in Fig.~\ref{fig:window_ablation} with different windowing around the predicted timestamp. In particular, we match the narration to video content that comes \emph{before} the predicted start and end times. This is to test whether video and narration are reasonably aligned, or if the narration captures actions that took place much earlier in the video. We observe that different windowing strategies around the predicted timestamps make a minimal difference and all perform better than the approach without precise localization described in Sec. \ref{sec:prec-loc-ablation} on most metrics.

\input{figures/window_ablation}

\input{tables/order_ablation}

\subsection{Does the order of training on auxiliary data affect generalization?} 

Next, we include ablations on the order of training on the auxiliary multimodal data and each stage of the data processing pipeline. 

In Tab.~\ref{tab:order_ablation}, we compare two strategies for incorporating multiple data sources: sequential training and joint training. In the sequential setup, we first train on text-only data, followed by fine-tuning on video-text (commentary + OOD) data. In the joint setup, all auxiliary data sources and OOD feedback data are combined and trained on simultaneously. We find that sequential training slightly outperforms joint training across all metrics, suggesting that first learning the domain-relevant language prior to learning video-conditional generation may facilitate better generalization. However, the differences are relatively small, indicating that both strategies are viable for leveraging auxiliary data.

\input{tables/filtering_ablation}

\subsection{Does our pipeline to improve the quality of the competition data improve generalization ability?}
\label{sec:prec-loc-ablation}
We validate each design decision of our pipeline: training with OOD feedback data, refining the in-the-wild commentary data, and performing precise localization. To test the impact of skipping precise localization without decreasing the sampling rate or excessive memory usage, we produce 4-second clips by running a sliding window through the ASR segment. Each clip is paired with the refined commentary from its corresponding ASR segment. 
 
 In Tab.~\ref{tab:ablation_filtering}, we see that using the noisy in-the-wild commentary data alone yields worse performance compared to zero-shot performance in Tab. 1 in the main paper. Combining with OOD feedback data improves performance, however, the most significant improvements come from refinement and the precise localization stages of the data processing pipeline.

\subsection{LLM refinement} 

The supplementary file also provides all prompt templates and annotation references used in our study. These include prompt instructions provided to LLMs for ASR commentary refinement, precise localization of refined commentary, and rating of feedback specificity and actionability. 

To refine noisy ASR outputs from competition videos into concise and anonymized commentary, we prompt an LLM to remove irrelevant details and retain key information related to pose, body movement, and quality of execution. The prompt below was used to guide this refinement process. 

\begin{tcolorbox}[promptbox, title=LLM Refinement Prompt]
You are an ASR refiner for a rock climbing competition. You take ASR inputs which may be noisy and not concise and output concise commentary. Replace named entities with general terms (e.g., “person”, “competition”). Focus on capturing the action, body parts, pose information, and quality of movement. If the input is unintelligible or contains only music or applause, return only: \texttt{[SKIP]}.

\textbf{Input:}

\texttt{ASR narration: \{narration\}}

\textbf{Output:}

Cleaned narration:
\end{tcolorbox}

\subsection{Precise localization of refined commentaries} To enable more precise timestamps of refined commentary in ASR transcripts, we prompt an LLM to localize each commentary segment to a short (1–4 second) time span based on context provided by the word-level timestamps. See prompt below.

\begin{tcolorbox}[promptbox, title=Precise localization of refined commentaries Prompt]
You are an ASR refiner for rock climbing competition commentary. Your task is to match cleaned, anonymized commentary with corresponding timestamps in noisy ASR input. The ASR input consists of transcriptions with word-level timestamps. Your goal is to determine when each action in the cleaned commentary occurs in the ASR by finding the most relevant timestamps. Each action should be localized to a \textbf{1s–4s time span} based on matching words and phrases. If an exact match is unavailable, use the closest approximation. Do \textbf{not} return timestamps for unintelligible sections (music, applause, background noise). Ensure all timestamps are precise and correspond to the moment when the action is happening.

Format the output as a structured list where each commentary line is paired with its estimated time range (corresponding \textbf{1s–4s} range) in the ASR. Please be concise.

\textit{Example output format:}
\begin{quote}

[
  {"commentary": "The climber hooks the toe on the right and pulls himself up.", "timestamp": (47.8, 49.66)},
  {"commentary": "He reaches for the next hold.", "timestamp": (50.5, 52.3)}
]

\end{quote}

\textbf{Return only the JSON list}—no other explanations, markdown, or formatting characters.

\textbf{ASR with word-level timestamps:} \texttt{\{whisper\_transcript\}}\\
\textbf{Refined commentary:} \texttt{\{refined\_commentary\}}

\textbf{Output:}
\end{tcolorbox}

\subsection{Automatic evaluation prompts} We assess the specificity and actionability of generated feedback. This LLM-based rating helps us evaluate feedback quality more interpretably. See the prompts below.

Table~\ref{tab:specificity_table} presents the full set of specificity examples (levels 1–4) shown to annotators. These show how feedback becomes more informative and elaborative as specificity increases. Similarly, Table~\ref{tab:actionability_table} shows feedback examples at each level of the actionability scale (1–3). Both of these tables were used to train annotators and calibrate (via in-context learning) LLM scoring models.

\begin{tcolorbox}[promptbox, title=Specificity Rating Prompt]
Analyze the given generated feedback and provide only a numerical rating for the **generated feedback**, from \textbf{1 to 4}, where 1 means "not specific" and 4 means "very specific".

\textbf{Definition:} Feedback conveys details about current movement and corrective measures. Specificity refers to the precision of movement information, focusing on the present; actionability guides future adjustments.

\textbf{Ratings Guide:}
\begin{itemize}
    \item \textbf{1 – Least Specific:} Very vague, offers little useful information.
    \item \textbf{2 – Vague:} Mentions either movement pattern details \textit{or} quality descriptors (e.g., smoothness, stiffness), or just performance outcomes.
    \item \textbf{3 – Slightly Specific:} Connects movement details to quality indicators but lacks elaboration.
    \item \textbf{4 – Very Specific:} Precise movement and quality info with elaboration (e.g., when, why, or how).
\end{itemize}

\textbf{Examples:}
\begin{Verbatim}[breaklines=true, breakanywhere=true]
Rating '1':
  - "The shot could be improved."
Rating '2':
  - "The shooter is standing up straight"
Rating '3':
  - "Standing straight up limits explosiveness and lift"
Rating '4':
  - "Standing straight up limits explosiveness and lift because it prevents your lower body from fully loading the muscles needed for an explosive push-off."

Rating '1':
  - "The climber needs to have more confidence."
Rating '2':
  - "The climber hesitates before reaching for the higher hold"
Rating '3':
  - "The climber hesitates and takes a shorter step when reaching for the higher hold."
Rating '4':
  - "The climber hesitates and takes a shorter step when reaching for the higher hold, which limits the momentum needed to successfully grab it."

Rating '1':
  - "The player is dribbling poorly"
Rating '2':
  - "The contact with the ball is closer to the heel rather than through that inside curvature of the foot. "
Rating '3':
  - "The contact with the ball is closer to the heel rather than through that inside curvature of the foot which affects controllability"
Rating '4':
  - "The contact with the ball is closer to the heel...so we have less control over the direction of the pass."
\end{Verbatim}
\end{tcolorbox}

\begin{tcolorbox}[promptbox, title=Actionability Rating Prompt]
Analyze the generated feedback and provide only a numerical rating for the \textbf{generated feedback}, from \textbf{1 to 3}, where 1 means "not actionable" and 3 means "actionable".

\textbf{Definition:} Actionability refers to the degree to which feedback can be implemented by the learner (e.g., specific corrective directions). This scale evaluates how directly feedback helps guide performance adjustments.

Skip scoring if the feedback is purely positive reinforcement.

\textbf{Scale:}
\begin{itemize}
    \item \textbf{Skipped} – If the feedback is only positive reinforcement.
    \item \textbf{1 – Not Actionable:} Vague or lacks any clear guidance the learner can act on.
    \item \textbf{2 – Minimally Actionable:} Identifies what to change, but not how to do it.
    \item \textbf{3 – Actionable:} Provides specific, clear directions to help the learner adjust.
\end{itemize}

\textbf{Example Progressions:}
\begin{Verbatim}[breaklines=true, breakanywhere=true]
Rating '1':
- "That wasn’t quite right."
Rating '2':
- "Your stance is off-balance."
Rating '3':
- "Widen your stance to be shoulder-length apart and keep your weight centered over your feet to maintain balance."

Rating '1':
- "The climber could use a more efficient technique."
Rating '2':
- "The climber is using a one-hand hold start, which is a good technique for beginners, but may not be the most efficient for experienced climbers."
Rating '3':
- "For a more efficient climb, try switching from a one-hand hold start to a two-handed start and engage both your hands and core simultaneously so you can distribute your weight evenly."

Rating '1':
- "Your form is poor."
Rating '2':
- "Your arm bent too much."
Rating '3':
- "Keep your arm straight until you initiate the follow-through."

Rating '1':
- "The player is dribbling poorly."
Rating '2':
- "The player’s dribble lacks control because their touches are inconsistent."
Rating '3':
- "Use smaller, more controlled touches on the ball and stay on the balls of your feet to maintain better control."

Rating '1':
- "The player’s first touch was off."
Rating '2':
- "The first touch is slow and takes them in the wrong direction."
Rating '3':
- "The player's first touch is slow and takes them in the wrong direction, causing them to take an extra touch and lose time. They need to move their feet in the direction they want to go."

Rating '1':
- "The ball’s trajectory was off."
Rating '2':
- "The ball’s trajectory is too flat."
Rating '3':
- "Release the ball slightly earlier and follow through higher to create a better arc on your shot."

Rating '1':
- "The climber is struggling."
Rating '2':
- "The climber should work on improving their grip strength and endurance through training."
Rating '3':
- "The climber could improve their grip strength and endurance by incorporating more finger exercises and grip strengthening exercises into their training routine."

Rating '2':
- "Could improve by keeping their feet closer together and using their hips to generate power."
Rating '3':
- "Improve by adjusting foot positioning and engaging the hips more."
\end{Verbatim}
\end{tcolorbox}

\subsection{Instructions for annotators}

We include tables showing example feedback at each specificity and actionability level, which were provided to annotators. 

\onecolumn
\begin{table*}[t]
\centering
\begin{tabular}{|p{0.23\linewidth}|p{0.23\linewidth}|p{0.23\linewidth}|p{0.23\linewidth}|}
\hline
\textbf{Level 1 (Least Specific)} & \textbf{Level 2 (Vague)} & \textbf{Level 3 (Slightly Specific)} & \textbf{Level 4 (Very Specific)} \\
\hline
The shot could be improved. & The shooter is standing up straight. & Standing straight up limits explosiveness and lift. & Standing straight up limits explosiveness and lift because it prevents your lower body from fully loading the muscles needed for an explosive push-off. \\
\hline
The shot is poor. & Your arm was bent too much. & Your arm was bent too much causing the shot to look stiff. & Your guide arm was bent too much prior to lifting up to the release point, and caused the shot to look stiff. \\
\hline
The player missed the shot. & The ball’s trajectory is flat. & The ball’s trajectory is flat because the release point is too late. & The ball’s trajectory is flat because the release point is too late. This is because the shoulders and hips are slow to rotate. \\
\hline
The climber needs to have more confidence. & The climber hesitates before reaching for the higher hold. & The climber hesitates and takes a shorter step when reaching for the higher hold. & The climber hesitates and takes a shorter step when reaching for the higher hold, which limits the momentum needed to successfully grab it. \\
\hline
The woman is doing a good job climbing the wall. & The climber's movements are smooth and controlled. & The climber is executing a great job, with a smooth and controlled movement, especially in transitions between holds. & The climber is executing a great job, with a smooth and controlled movement due to excellent foot placement and efficient weight transfer, especially in transitions between holds. \\
\hline
The climber has good technique. & The climber maintains good control. & The climber has successfully placed their right foot on a ledge and released their left foot to add force which helps with control. & The climber has successfully placed their right foot on a ledge, applying sufficient pressure, and released their left foot to add force to the right foot, which will help them stay pulled into the wall leading to more control. \\
\hline
\end{tabular}
\caption{Examples of feedback at different specificity levels (1–4).}
\label{tab:specificity_table}

\end{table*}
\twocolumn

\onecolumn
\begin{table*}[t]
\centering
\begin{tabular}{|p{0.3\linewidth}|p{0.3\linewidth}|p{0.3\linewidth}|}
\hline
\textbf{Level 1 (Not Actionable)} & \textbf{Level 2 (Minimally Actionable)} & \textbf{Level 3 (Actionable)} \\
\hline
That wasn’t quite right. & Your stance is off-balance. & Widen your stance to be shoulder-length apart and keep your weight centered over your feet to maintain balance. \\
\hline
The climber could use a more efficient technique. & The climber is using a one-hand hold start, which is a good technique for beginners, but may not be the most efficient for experienced climbers. & For a more efficient climb, try switching from a one-hand hold start to a two-handed start and engage both your hands and core simultaneously so you can distribute your weight evenly. \\
\hline
Your form is poor. & Your arm bent too much. & Keep your arm straight until you initiate the follow-through. \\
\hline
The player is dribbling poorly. & The player’s dribble lacks control because their touches are inconsistent. & Use smaller, more controlled touches on the ball and stay on the balls of your feet to maintain better control. \\
\hline
The player’s first touch was off. & The first touch is slow and takes them in the wrong direction. & The player's first touch is slow and takes them in the wrong direction, causing them to take an extra touch and lose time. They need to move their feet in the direction they want to go. \\
\hline
The ball’s trajectory was off. & The ball’s trajectory is too flat. & Release the ball slightly earlier and follow through higher to create a better arc on your shot. \\
\hline
The climber is struggling. & The climber should work on improving their grip strength and endurance through training. & The climber could improve their grip strength and endurance by incorporating more finger exercises and grip strengthening exercises into their training routine. \\
\hline
-- & Could improve by keeping their feet closer together and using their hips to generate power. & Improve by adjusting foot positioning and engaging the hips more. \\
\hline
\end{tabular}
\caption{Examples of feedback at different actionability levels (1–3).}
\label{tab:actionability_table}
\end{table*}
\twocolumn

\subsection{LLM biases} LLMs are known to carry various biases. This can be problematic when using LLMs as an evaluator. To explore these biases within the context of using an LLM to automatically score specificity and actionability, we consider two types of bias: gender and length bias. 
\\\indent To assess gender bias, we rewrite a set of feedback with male and female pronouns and compute specificity and actionability scores for each. First, we randomly sampled 20 feedbacks from the ExpertAF dataset. For each of the 20 examples, we manually created a version that uses male pronouns and another that uses female pronouns. We then calculated the specificity and actionability scores using GPT4o. There were no differences in the actionability and specificity scores. This may indicate that biases from LLMs may occur in open-ended generation and more abstract ratings such as sentiment, in contrast, specificity and actionability judge the structure and the definitions are not dependent on gendered pronouns.
\\\indent To assess length bias, we report the effect of adding a neutral phrase to increase the feedback length. We append a neutral phrase (i.e. ``This was observed during practice'') that should not impact specificity or actionability. We observe a 0.05 increase in specificity (2.95 to 3.00) and actionability (2.2 to 2.25) when the neutral text is appended to the feedback. This indicates a slight bias to longer outputs, however, the difference is still small.

%% file: figures/window_ablation.tex
\begin{figure*}
    \centering
    \includegraphics[width=\linewidth]{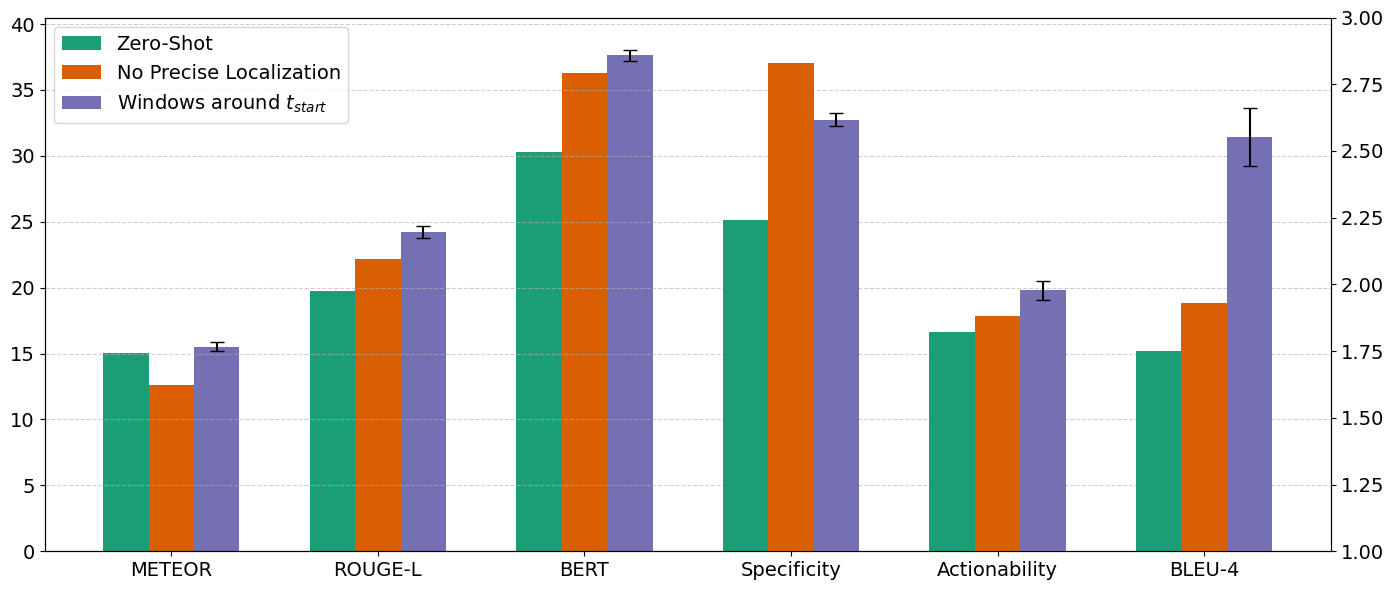}
    \caption{\textbf{Window ablation.} $t_{\text{start}}$ and $t_{\text{end}}$ are the start and end timestamps produced by the precise localization step. For \textit{Windows around $t_{\text{start}}$}, the mean and confidence interval are computed over the performance of the following windowing strategies when used for training: $(t_{\text{start}}, t_{\text{end}})$, $(t_{\text{start}}-3, t_{\text{start}}+1)$, $(t_{\text{start}}-4, t_{\text{start}})$, and $(t_{\text{start}}-4, t_{\text{end}})$. These windows experiment with different ways of including actions that may have occurred prior to the narration. Observe that the confidence interval is very small, indicating comparable performance when the window is slightly shifted earlier, but consistent improvement over lack of precise localization.}
    \label{fig:window_ablation}
\end{figure*}

%% file: tables/order_ablation.tex
\begin{table}[h]
    \centering
    \begin{tabular}{l|ccc}
    \hline
     Order & \textbf{BLEU-4} & \textbf{METEOR} & \textbf{ROUGE-L} \\
    \hline
      Sequential & \textbf{2.68} $\pm$ 0.02 & \textbf{15.59} $\pm$ 0.14 & \textbf{24.01} $\pm$ 0.05 \\
      Joint & 2.63 ± 0.03	& 15.47 ± 0.04 & 23.87 ± 0.04 \\
    \end{tabular}
    \caption{Training order ablation. Sequential is training on text first, then on the video-text data. Joint is training on both data types together.}
    \label{tab:order_ablation}
\end{table}

%% file: tables/filtering_ablation.tex
\begin{table*}[ht]
\small
\centering
\begin{tabular}{ccccccc}
\hline
\textbf{OOD Fd.} & \textbf{R} & \textbf{PL} & \textbf{BLEU-4} & \textbf{METEOR} & \textbf{ROUGE-L} & \textbf{BERT} \\
\hline
\ding{55} & \ding{55} & \ding{55} & 1.06 ± 0.02 & 7.50 ± 0.06 & 13.65 ± 0.14 & 15.6 \\
\checkmark & \ding{55} & \ding{55} & 1.62 ± 0.04 & 13.44 ± 0.05 & 19.20 ± 0.06 & 29.1 \\
\checkmark & \checkmark & \ding{55} & 1.93 ± 0.01 & 12.60 ± 0.07 & 22.18 ± 0.10 & 36.3  \\
\checkmark & \checkmark & \checkmark & 2.67 ± 0.02 & 15.38 ± 0.10 & 23.39 ± 0.06 & 37.0  \\ 
\hline
\end{tabular}
\caption{Ablation of each stage of our commentary data processing pipeline and effect of adding OOD feedback from the source domain. Significant improvements from using the commentary data are from refinement and precise localization. R=refinement. PL=precise localization.}
\label{tab:ablation_filtering}
\end{table*}